\pdfoutput=1

\documentclass[11pt]{article}

\usepackage{EMNLP2023}

\usepackage{times}
\usepackage{latexsym}
\usepackage{graphicx}
\usepackage{caption}
\usepackage{subcaption}
\usepackage{url}
\usepackage{enumitem}
\usepackage{multirow}
\usepackage{hhline}
\usepackage{multicol}
\usepackage{amssymb}
\usepackage{tablefootnote}

\usepackage[T1]{fontenc}

\usepackage[utf8]{inputenc}

\usepackage{microtype}

\usepackage{inconsolata}

\newcommand\blfootnote[1]{%
  \begingroup
  \renewcommand\thefootnote{}\footnote{#1}%
  \addtocounter{footnote}{-1}%
  \endgroup
}

\title{Theory of Mind in Large Language Models: Examining Performance of 11 State-of-the-Art models vs. Children Aged 7-10 on Advanced Tests}

\author{Max van Duijn\textsuperscript{1}*, Bram van Dijk\textsuperscript{1}*, Tom Kouwenhoven\textsuperscript{1}*, \\ {\bf Werner de Valk\textsuperscript{1}}, {\bf Marco Spruit\textsuperscript{1,}\textsuperscript{2}}, \and {\bf Peter van der Putten\textsuperscript{1}}\\
 \textsuperscript{1}Leiden Institute of
  Advanced Computer Science \\
  \textsuperscript{2}Leiden University Medical Centre \\
  Corresponding author: \texttt{m.j.van.duijn@liacs.leidenuniv.nl}}

\begin{document}
\maketitle
\begin{abstract}
To what degree should we ascribe cognitive capacities to Large Language Models (LLMs), such as the ability to reason about intentions and beliefs known as Theory of Mind (ToM)? Here we add to this emerging debate by (i) testing 11 base- and instruction-tuned LLMs on capabilities relevant to ToM beyond the dominant false-belief paradigm, including non-literal language usage and recursive intentionality; (ii) using newly rewritten versions of standardized tests to gauge LLMs' robustness; (iii) prompting and scoring for open besides closed questions; and (iv) benchmarking LLM performance against that of children aged 7-10 on the same tasks. We find that instruction-tuned LLMs from the GPT family outperform other models, and often also children. Base-LLMs are mostly unable to solve ToM tasks, even with specialized prompting. We suggest that the interlinked evolution and development of language and ToM may help explain what instruction-tuning adds: rewarding cooperative communication that takes into account interlocutor and context. We conclude by arguing for a nuanced perspective on ToM in LLMs.\blfootnote{*Equal contribution.}
\end{abstract}

\section{Introduction}\label{Intro}
Machines that can think like us have always triggered our imagination. Contemplation of such machines can be traced as far back as antiquity \citep{livelythomas2020homer}, and peaked with the advent of all kinds of `automata' in the early days of the Industrial Revolution \citep{voskuhl2013one} before settling in computer science from the 1950s \citep{turing1950comp}. Currently people around the world can interact with powerful chatbots driven by Large Language Models (LLMs), such as OpenAI's ChatGPT \citep{openai2023gpt4}, and wonder to what degree such systems are capable of thought.

LLMs are large-scale deep neural networks, trained on massive amounts of text from the web. They are vastly complex systems: even if all details about their architecture, training data, and optional fine-tuning procedures are known (which is currently not the case for the most competitive models), it is very difficult to oversee their capabilities and predict how they will perform on a variety of tasks. Researchers from linguistics \cite{manning2020emergent}, psychology \citep{binz2023using, kosinski2023theory, webb2023emergent}, psychiatry \citep{kjell2023ai}, epistemology \citep{sileo2023mindgames}, logic \citep{creswell2022selection}, and other fields, have therefore started to study LLMs as new, `alien' entities, with their own sort of intelligence, that needs to be probed with experiments, an endeavour recently described as `machine psychology' \cite{hagendorff2023machine}. This not only yields knowledge about what LLMs are capable of, but also provides a unique opportunity to shed new light on questions surrounding our own intelligence \citep{dillion2023canAI, binz2023turning}.

Here we focus on attempts to determine to what degree LLMs demonstrate a capacity for Theory of Mind (ToM), defined as the ability to work with beliefs, intentions, desires, and other mental states, to anticipate and explain behaviour in social settings \cite{apperly2010mindreaders}. We first address the question \textbf{how LLMs perform} on standardized, language-based tasks used to assess ToM capabilities in humans. We extend existing work in this area, surveyed in Section \ref{background}, in four ways: by (i) testing 11 models (see Table \ref{tab:1}) for a broader suite of capabilities relevant to ToM beyond just the dominant false-belief paradigm, including non-literal language understanding and recursive intentionality (A \textit{wants} B to \textit{believe} that C \textit{intends}...); (ii) using newly written versions of standardized tests with varying degrees of deviation from the originals; (iii) including open questions besides closed ones; and (iv) benchmarking LLM performance against that of children aged 7-8 (n=37) and 9-10 (n=36) on the same tasks. Section \ref{methods} contains details of our test procedures for both children and LLMs. After reporting the results in Section \ref{results}, we turn to the question \textbf{how variation in performance of the LLMs we tested can be explained} in Section \ref{discussion}. We conclude by placing our findings in the broader context of strong links between language and ToM in human development and evolution, and tentatively interpret what it means for an LLM to pass (or fail) ToM tests.

We are aware of issues regarding LLM training and deployment, for example regarding the biases they inherit \cite{lucy-bamman-2021-gender, bender2021dangers}, problems for educators \cite{sparrow2022full-on}, and ethical concerns in obtaining human feedback \cite{perrigo2023openai}. Ongoing reflection on the use of LLMs is necessary, but outside the scope of this paper. 

\section{Background}\label{background}

\subsection{Large Language Models}
The field of Natural Language Processing (NLP) has been revolutionized by the advent of Transformer models \citep{vaswani2017attention, devlin2019bert}, deep neural networks that can induce language structures through self-supervised learning. During training, such models iteratively predict masked words from context in large sets of natural language data. They improve at this task by building representations of the many morphological, lexical, and syntactic rules governing human language production and understanding \citep{manning2020emergent, rogers2020primer, grand2022semantic}. Models exclusively trained through such self-supervision constitute what we refer to as `base-LLMs' in this paper.

Base-LLMs can generate natural language when prompted with completion queries (`A mouse is an ...'). They can also be leveraged successfully for an array of other challenges, such as question-answering and translation, which often requires task-specific fine-tuning or prompting with specific examples, known as few-shot-learning \citep{brown2020language}. This makes them different from a new generation of LLMs that we refer to as `instruct-LLMs' in this paper, and to which the currently most competitive models belong. In instruction-tuning, various forms of human feedback are collected, such as ranking most suitable responses, which then forms the reward-signal for further aligning these models to human preferences through reinforcement learning \citep{ouyang2022training}. The resulting LLMs can be prompted with natural language in the form of instructions to perform a wide variety of tasks directly, amounting to zero-shot learning \citep{wei2022finetuned}.

A key realization is thus that LLMs are given either no explicitly labelled data at all, or, in the case of instruct-LLMs, data with human labels pertaining to relatively general aspects of communicative interaction. As such they are part of a completely different paradigm than earlier language models that were trained on, for example, data sets of human-annotated language structures \citep[e.g.][]{nivre2016universal}. This means that when LLMs are capable of such tasks as solving co-reference relationships or identifying word classes \citep{manning2020emergent}, this arises as an \textit{emergent} property of the model's architecture and training on different objectives. Given that such emergent linguistic capabilities have been observed \citep{reif2019visualizing, grand2022semantic}, it is a legitimate empirical question which other capacities LLMs may have acquired as `by-catch'.

\subsection{Theory of Mind in Humans and LLMs}
ToM, also known as `mindreading', is classically defined as the capacity to attribute mental states to others (and oneself), in order to explain and anticipate behaviour. The concept goes back to research in ethology in which \citet{premack1978does} famously studied chimpanzees' abilities to anticipate behaviour of caretakers. When focus shifted to ToM in humans, tests were developed that present a scenario in which a character behaves according to its \textit{false beliefs} about a situation, and not according to the reality of the situation itself––which a successful participant, having the benefit of spectator-sight, can work out (see Section \ref{tomtests}). 

Initial consensus that children could pass versions of this test from the age of 4 was followed by scepticism about additional abilities it presumed, including language skills and executive functioning, which led to the development of simplified false-belief tests based on eye-gaze that even 15 month-olds were found to `pass' \citep{onishi2005do15}. While this line of research also met important criticism \citep[for a review see][]{barone2019infants}, it highlights two key distinctions in debate from the past decades: implicit-behavioural versus explicit-representational and innate versus learned components of ToM. Some researchers see results from eye-gaze paradigms as evidence for a native or very early developing capacity for belief-attribution in humans \citep{carruthers2013mindreading} and hold that performance on more complex tests is initially `masked' by a lack of expressive skills \citep[cf. also][]{fodor1992theory}. Others have attempted to explain eye-gaze results in terms of lower-level cognitive mechanisms \citep{heyes2014false} and argued that the capacity for belief-attribution itself develops gradually in interaction with more general social, linguistic, and narrative competencies \citep{heyesfrith2014cultural, milligan2007language, hutto2008folk}. Two-systems approaches \citep{apperly2010mindreaders} essentially reconcile both sides by positing that our mindreading capacity encompasses both a basic, fast, and early developing component and a more advanced and flexible component that develops later.

In computational cognitive research, a variety of approaches to modelling ToM have been proposed \citep[e.g.][]{baker2011bayesian, arslan2017five}. More recently neural agents \citep{pmlr-v80-rabinowitz18a} have been implemented, along with an increasing number of deep-learning paradigms aimed at testing first- and second-order ToM via question-answering. Initially this was done with recurrent memory networks \cite{grant2017can, nematzadeh-etal-2018-evaluating} using data sets of classic false-belief tests from psychology, but after issues surfaced with simple heuristics for solving such tasks, scenarios were made more varied and challenging \cite{le-etal-2019-revisiting}. From the inception of BERT as one of the first LLMs \cite{devlin2019bert}, we have seen roughly two approaches for testing ToM in LLMs: many different ToM scenarios integrated in large benchmark suites \citep[e.g.][]{sap-etal-2022-neural, srivastava2023beyond, sileo2023mindgames, ma2023tomchallenges, shapira2023clever}, and studies that modified standardized ToM tests as used in developmental and clinical research for prompting LLMs \citep[e.g.][]{kosinski2023theory, ullman2023large, bubeck2023sparks, brunet2023conversational, chowdhery2022palm, moghaddam2023boosting, marchetti2023developing}. This paper adds to the latter tradition in four respects, as listed in the introduction.

\section{Methodology}\label{methods}
Here we describe our tasks and procedures for testing LLMs and children; all code, materials, and data are on OSF: \url{https://shorturl.at/FQR34}.

\subsection{ToM Tests}\label{tomtests}

\textbf{Sally-Anne test, first-order (SA1)} –– 
The Sally-Anne test \cite{wimmer1983beliefs, baron1985} is a classic first-order false belief test. It relies on a narrative in which Sally and Anne stand behind a table with a box and a basket on it. When Anne is still present, Sally puts a ball in her box. When Sally leaves, Anne retrieves the ball from the box and puts it in her own basket. The story ends when Sally returns and the participant is asked the experimental question `Where will Sally look for the ball?' The correct answer is that she will look in her box. We followed up by asking a motivation question, `Why?', to prompt an explanation to the effect of `she (falsely) believes the object is where she left it'. 

\textbf{Sally-Anne test, second-order (SA2)} –– 
While SA1 targets the participant's judgement of what a character \textit{believes} about the location of an unexpectedly displaced object, in SA2 the participant needs to judge what a character \textit{believes} that \textit{another character believes} about the location of an ice-cream truck \citep{perner1985}. Sally and Anne are in a park this time, where an ice-cream man is positioned next to the fountain. Anne runs home to get her wallet just while the ice-cream man decides to move his truck to the swings. He tells Sally about this, but unknown to her, he meets Anne on the way and tells her too. Sally then runs after Anne, and finds her mother at home, who says that Anne picked up the wallet and went to buy ice cream. The experimental question now is `Where does Sally think Anne went to buy ice cream?', with as correct answer `to the fountain', also followed up with `Why?', to prompt an explanation to the effect of `Sally doesn't know that the ice-cream man told Anne that he was moving to the swings'.

\textbf{Strange Stories test (SS)} ––
The Strange Stories test \cite{happe1994, kaland2005strange} depicts seven social situations with non-literal language use that can easily be misinterpreted, but causes no problems to typically developed adults. To understand the situations, subjects must infer the characters' intentions, applying ToM. For example, in one of the items a girl wants a rabbit for Christmas. When she opens her present, wrapped in a big enough box, it turns out that she received a pile of books. She says that she is really happy with her gift, after which subjects are asked the experimental question `Is what the girl says true?', with correct answer `No'. They can motivate their answer after the question `Why does she say this?', with as correct answer `to avoid her parents' feelings being hurt'. Items increase in difficulty and cover a lie, pretend-play scenario, practical joke, white lie (example above), misunderstanding, sarcasm, and double bluff.
 
\textbf{Imposing Memory test (IM)} ––
The Imposing Memory test was originally developed by \citet{kinderman1998imt}, but the test has been revised several times; we rely on an unpublished version created by Anneke Haddad and Robin Dunbar \cite{vanduijn2016}, originally for adolescents, which we adapted thoroughly to make it suitable for children aged 7-10. Our version features two different stories, followed by true/false questions, 10 of which are `intentionality' and 12 are `memory' questions. For instance, in one story Sam has just moved to a new town. He asks one of his new classmates, Helen, where he can buy post stamps for a birthday card for his granny. When Helen initially sends him to the wrong location, Sam wonders whether she was playing a prank on him or just got confused about the whereabouts of the shop herself. He goes and asks another classmate, Pete, for help. As in the original IM, the intentionality questions involve reasoning about different levels of recursively embedded mental states (e.g., at third-level: `Helen \textit{thought} Sam \textit{did not believe} that she \textit{knew} the location of the store that sells post stamps'), whereas the memory questions require just remembering facts presented in the story (e.g., to match third-level intentionality questions, three elements from the story are combined: `Sam was looking for a store where they sell post stamps. He told Pete that he had asked Helen about this'). 

\subsection{Scoring Test Answers}\label{scoringopenquestions}
Test scores for both children and LLMs were determined in the following way. For each of the SA1 and SA2 items, as well as for the seven SS items, a correct answer to the experimental question yielded 1 point. These answers were discrete and thus easy to assess (`box', `fountain', `no', etc.). For the motivation question a consensus score was obtained from two expert raters, on a range from 0-2, with 0 meaning a missing, irrelevant, or wrong motivation, 1 meaning a partly appropriate motivation, and 2 meaning a completely appropriate motivation that fully explained why the character in each scenario did or said something, or had a mental or emotional mind state. Thus, the maximum score for the SA1, SA2, and SS was 3 points per item, which were averaged to obtain a score between 0 and 1. For each correct answer to a true/false question in the IM, 1 point was given. All scores and ratings can be found on OSF.

\subsection{Deviations} \label{devs}
We tested the LLMs on the original SA and SS scenarios, but also on manually created \textit{deviations} that increasingly stray from their original formulations, to prevent LLMs from leveraging heuristics and memorizing relevant patterns from the training data. Thus, deviations probe the degree to which performance on ToM tests in LLMs generalizes. Deviation 0 was always the original test scenario (likely present in the training data); deviation 1 was a superficial variation on the original with only e.g., objects and names changed (similar to \citet{kosinski2023theory}), whereas deviation 2 was a completely new scenario where only the ToM-phenomenon at issue was kept constant (e.g. `second-order false belief' or `irony'). Since our adaptation of the IM test has hitherto not been used or published, we did not include deviations for this test. 

\subsection{Test Procedures for LLMs}\label{procedureLLMs}
We leveraged 11 state-of-the-art LLMs: 4 base-LLMs and 7 instruct-LLMs (see Table \ref{tab:1}). Inference parameters were set such that their output was as deterministic as possible (i.e. a temperature $\approxeq$ zero or zero where possible) improving reproducibility. Each inference was done independently to avoid in-context learning or memory leakage between questions. This means that for each question, the prompt repeated the following general structure: [\textit{instruction}] + [\textit{test scenario}] + [\textit{question}].

\begin{table}[t] \small
\begin{center}
\begin{tabular}{c  c  c }
    \hline
    \textbf{Base-LLMs} & \textbf{Source} & \textbf{Size} \\ [0.3ex]
    \hline
    Falcon & \citet{penedo2023refinedweb} & 7B \\
    LLaMA & \citet{touvron2023llama} & 30B \\
    GPT-davinci & \citet{brown2020language} & 175B \\
    BLOOM & \citet{scao2022bloom} & 176B \\
    \hline
     \textbf{Instruct-LLMs} & ''  & ''  \\
    \hline
    Falcon-instruct & \citet{penedo2023refinedweb} & 7B \\
    Flan-T5 & \citet{flanT5_2022paper} & 11B \\
    GPT-3 \\ (text-davinci-003) & \citet{ouyang2022training} & 175B \\
    GPT-3.5-turbo & \citet{ouyang2022training} & 175B \\
    PaLM2 & \citet{anil2023palm} & 175-340B \\
    PaLM2-chat & \citet{anil2023palm} & 175-340B \\
    GPT-4 & \citet{openai2023gpt4} & >340B \\
    \hline
\end{tabular}
\end{center}
\caption{LLMs used in this study. Model sizes are undisclosed for GPT-4 and for PaLM2 and PaLM2-chat, thus we base ourselves on secondary sources for estimations; \citet{wired2023gpt-4} and \citet{cnbc2023PaLM2}, respectively.}
\label{tab:1}
\end{table}

Instruct-LLMs were prompted in a question-answering format that stayed as close as possible to the questionnaires given to children, without any further custom prompting or provision of examples. Instructions were also similar to those given to children (e.g. `You will be asked a question. Please respond to it as accurately as possible without using many words.'). The `Why'-questions in SA1 and SA2 were created by inserting the experimental question and answer the LLM gave into the prompt: [\textit{instruction}] + [\textit{test scenario}] + [\textit{experimental question}] + [\textit{LLM answer}] +[\textit{`Why?'}]. This was not necessary for SS, given that experimental and motivation questions could be answered independently.

For base-LLMs, known to continue prompts rather than follow instructions, staying this close to the children's questionnaires was not feasible. For the SA and SS we therefore fed base-LLMs the scenario as described before, but formulated the questions as text-completion exercises (e.g. `Sally will look for the ball in the '). Additionally, when creating the motivation questions for SA1 and SA2, we inserted the \textit{correct} answer to the experimental question, instead of the LLM's answer. This was because base-LLMs so often derailed in their output that the method described for instruct-LLMs did not yield sensible prompts. Base-LLMs thus had an advantage here over children and instruct-LLMs, who were potentially providing a motivation following up on an incorrect answer they gave to the experimental question.

For the closed questions in the IM we attempted to streamline the output of base-LLMs by including two example continuations in the desired answer format. These examples were based on trivial information we added to the scenarios, unrelated to the actual experimental questions. For example: `Helen: I wear a blue jumper today. This is [incorrect]', where it was added in the story that Helen wears a green jumper. This pushed nearly all base-LLM responses towards starting with `[correct]' or `[incorrect]', which we then assessed as answers to the true/false questions. We considered a similar prompt structure for SA and SS, amounting to adopting few-shot learning for base-LLMs throughout \citep{brown2020language}, but given that reformulating questions as text-completion exercises was by itself effective to get the desired output format, we refrained from inserting further differences from how instruct-LLMs are prompted. It is important to note that our prompts were in general not optimized for maximal test performance, but rather designed to stay as uniform and close to the way children were tested as possible, enabling a fair comparison among LLMs and with child performance. 

\subsection{Test Procedures for Children} \label{childproced}
Children were recruited from one Dutch and one international school in the South-West of the Netherlands: 37 children in the younger group (7-8y) and 36 children in the older group (9-10y). Children were administered digital versions of the SA and SS for the younger group, and of the IM for the older group, which they completed individually on tablets or PCs equipped with a touch screen. Test scenarios and questions were presented in a self-paced text format and all SA and SS questions were followed by an open text field in which they had to type their answer. As the IM features long scenarios, voice-overs of the text were included to alleviate reading fatigue. Here children had to answer by pressing yes/no after each question. To reduce memory bottlenecks, accompanying drawings were inserted (see OSF) and navigating back and forth throughout the tests was enabled. Informed consent for each child was obtained from caretakers, and the study was approved by the Leiden University Science Ethics Committee (ref. no. 2021-18). Test answers were evaluated and scored parallel to the approach for LLMs (Section \ref{scoringopenquestions}).

\section{Results}\label{results}

\begin{figure*}
    \centering
    \includegraphics[width=0.9\linewidth]{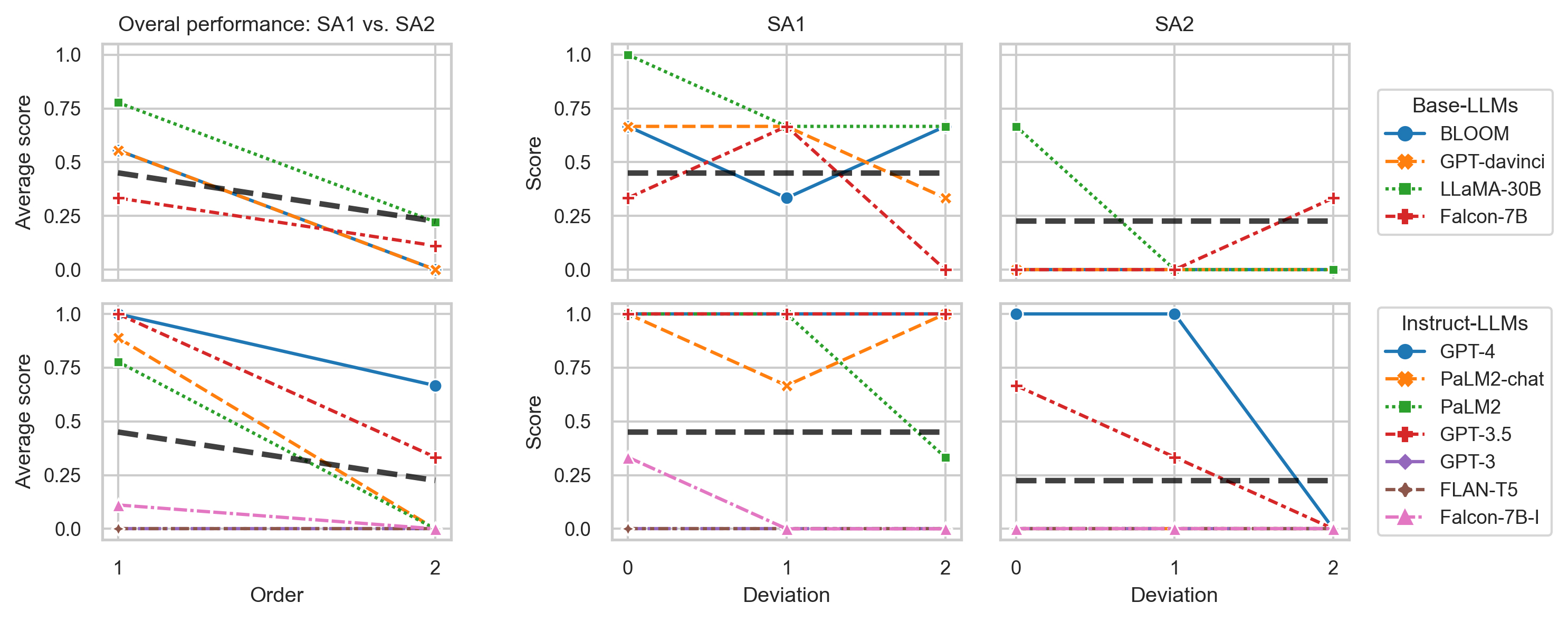}
    \caption{Performance on Sally-Anne tests for base-LLMs (top row) and instruct-LLMs (bottom row). Left column depicts performance on first- and second-order ToM (i.e. SA1 vs. SA2), averaged over the original and rewritten test versions. Middle and left columns depict performance for SA1 and SA2 over levels of deviation from the original test (0, 1, and 2; see Section \ref{devs}). Dashed lines indicate child performance (n=37, age 7-8 years).}
    \label{fig:1}
\end{figure*}

\subsection{Sally-Anne}
Overall performance on SA1 versus SA2 is given in Figure \ref{fig:1}, left column. Most base-LLMs perform above child level on first-order ToM (BLOOM, Davinci, LLaMA-30B) but fall at or or below child level on second-order ToM. A similar pattern is visible for instruct-LLMs: most models perform well above child level on first-order (GPT-4, GPT-3.5, PaLM2-chat, PaLM2), but not on second-order ToM. Exceptions are GPT-4 and GPT-3.5: while degrading on second-order, they remain above child level. For both base- and instruct-LLMs, smaller models tend to perform worse (Falcon-7B, Falcon-7B-I, FLAN-T5) with GPT-3's structurally low scores as striking exception. This is inconsistent with results reported by \citep{kosinski2023theory} for GPT-3, which is probably due to the fact that Kosinski applied a text-completion approach whereas we prompted GPT-3 with open questions.

When we consider the performance on SA1 and SA2 over deviations (middle and right columns in Figure \ref{fig:1}), we see once more that almost all LLMs struggle with second-order ToM, since performance decreases already on deviation 0 (i.e. the original test scenario), except for GPT-3.5 and GPT-4. Yet, it is the \textit{combination} of second-order ToM and deviation 2 that pushes also GPT-3.5 and GPT-4 substantially below child levels, except for Falcon-7B, although the chat-optimized version of this model (Falcon-7B-I) fails on all second-order questions. 

\begin{figure*}
\centering
\includegraphics[width=0.62\linewidth]{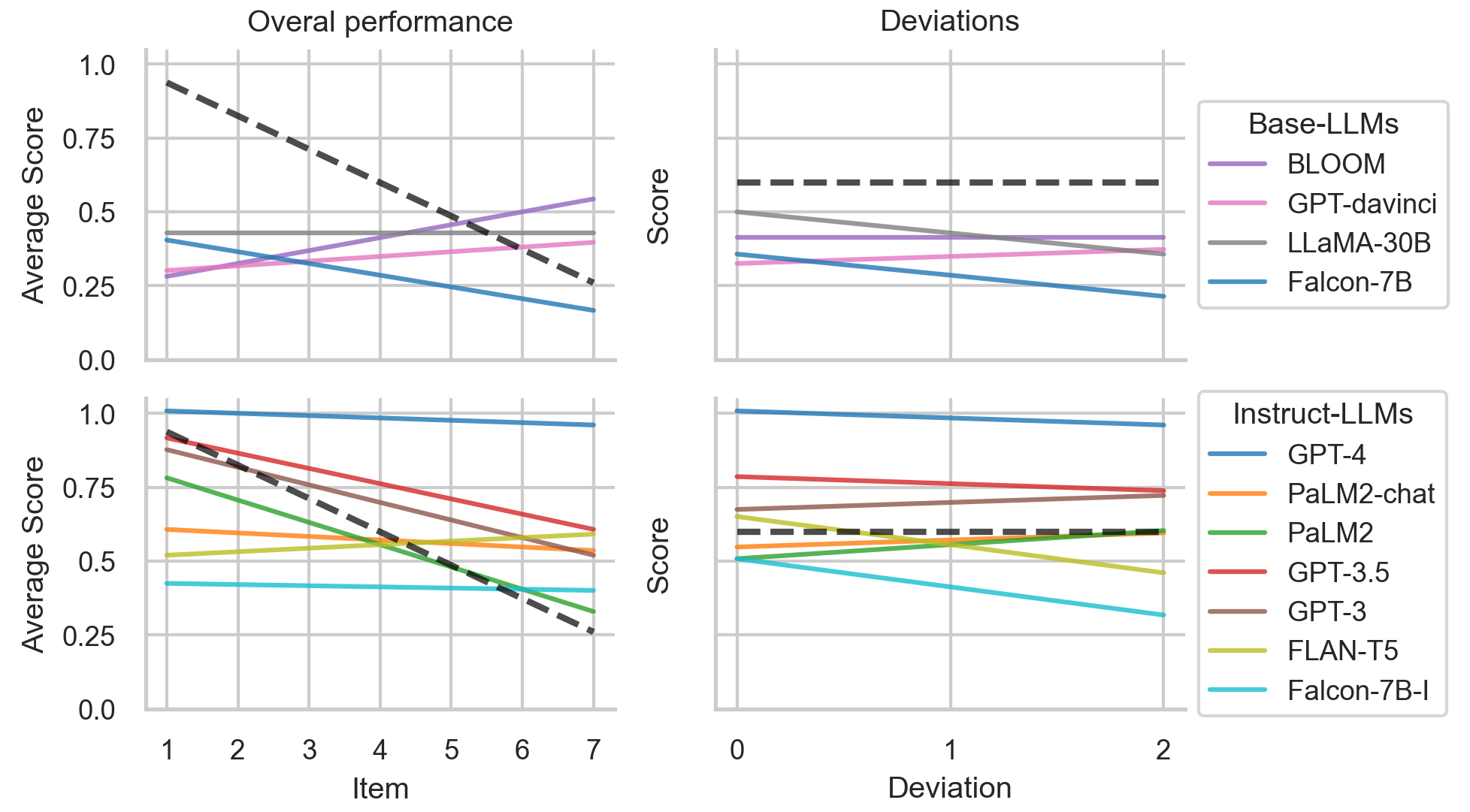}
\caption{Performance on Strange Stories for base-LLMs (top row) and instruct-LLMs (bottom row). Left column shows overall performance, averaged over levels of deviation from the original test. Right column shows performance over deviation levels, averaged over items. Dashed lines indicate child performance (n=37, 7-8y).}
\label{fig:2}
\end{figure*}

\subsection{Strange Stories}
General performance on SS is given in Figure \ref{fig:2}, left column. Whereas child performance declines as items become more complex (from 1 to 7; see Section \ref{tomtests}), this is overall less the case for LLM performance. As a result, all models surpass child level at some point, except for the smallest model, Falcon-7B. All base-LLMs score below child level on most items but perform above child level the most difficult ones, except Falcon-7B. For instruct-LLMs, we see that GPT-4 approaches perfect scores throughout. GPT-3 and GPT-3.5 perform at or close to child level on item 1, after which their performance somewhat declines, while staying well above child level. Other instruct-LLMs show a mixed picture: PaLM2-chat and FLAN-T5 surpass child level earlier than PaLM2. Interestingly, smaller FLAN-T5 outperforms large PaLM and PaLM2-chat on more difficult items. Falcon-7B-I, as smallest instruct-LLM, performs overall worst. 

If performance is plotted over deviations (right column in Figure \ref{fig:2}) we see little impact on most base-LLMs. For instruct-LLMs, it is striking that deviation levels have almost no effect on the larger models (GPT-4, PaLM2, PaLM2-chat, GPT-3, GPT-3.5), but do more dramatically lower performance of smaller models (FLAN-T5, Falcon-7B-I). In sum, base-LLMs perform below child level, except for the most complex items. Several large instruct-LLMs match or surpass child level throughout, others only for more complex items. Unlike for SA, deviation levels seem to have little negative impact. 

\subsection{Imposing Memory}\label{IMres}
The classical finding for the IM test is that error rates go up significantly for questions involving higher levels of recursive intentionality, but not for memory questions on matched levels of complexity, suggesting a limit to the capacity for recursive ToM specifically \citep{stiller2007perspective}.\footnote{While there is consensus in the literature that higher levels of intentionality are significantly harder for participants than lower levels, by various measures, there is debate about the difference with memory questions; see e.g. \citet{lewis2017higher}. For a critical discussion of measuring recursive intentionality in general, see \citet{wilson2023recursive}.} We verified this for our child data (n=36) with two mixed linear models for memory and intentional questions with random intercepts. We included five predictors that were contrast-coded such that each predictor indicated the difference in average performance with the previous level. For intentional questions, only the difference between level two and one was significant ($\beta=-0.222, p <.05$), marking a cut-off point after which performance remained consistently low. For memory questions, performance remained high across all levels ($>.85$), except for level four, where scores were significantly lower than at level three ($\beta=-0.292, p <.00$), but went up again at level five ($\beta=0.208, p <.00$). Thus, in line with earlier work, we find a cut-off point after which scores on intentionality questions remained consistently low, compared to scores on matched memory questions. We have no clear explanation for the dip in performance on memory questions at level four, but observe that it is driven by low scores on only one specific question out of a total of four for this level, which children may have found confusing. 

\begin{figure*}
    \centering
    \includegraphics[width=0.85\textwidth]{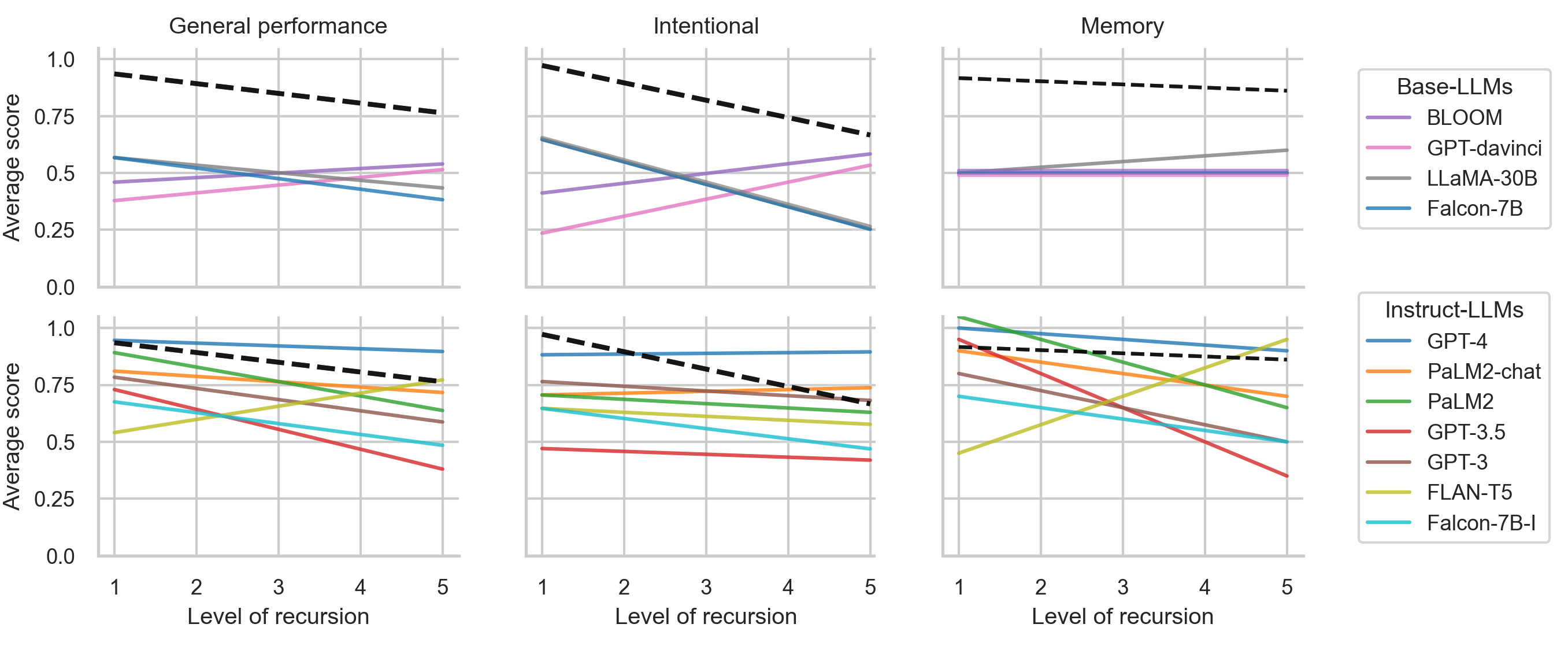}
\caption{Performance on Imposing Memory test for base-LLMs (top row) and instruct-LLMs (bottom row). Left column depicts overall performance over five levels of recursion, averaged over deviations. Middle and left columns depict performance for Memory and Intentional questions. Dashed lines indicate child performance (n=36, 9-10y).}
\label{fig:3}
\end{figure*}

In Figure \ref{fig:3} we see that all base-LLMs perform below child level, in general and on both intentionality and memory questions, and there is little variation in performance, except that larger base-LLMs (BLOOM, GPT-davinci) improve on higher levels of recursion. Regarding instruct-LLMs, we see largely the same picture, as they almost all perform below child level, in general and on both types of questions. The exception is GPT-4, which performs consistently well on all levels and stays above child level after second-order intentionality. For the difference between memory and intentional questions, instruct-LLMs perform better on easier memory questions, and drop towards the end, while on intentional questions, they already start lower and stay relatively constant. Lastly, it is remarkable that FLAN-T5, as one of the smallest instruct-LLMs, overall increases performance as recursion levels go up, and ends at child level. For GPT-3.5, which performs worst of all instruct-LLMs on this task, we see the exact opposite.

\subsection{Notes on Child Performance}
It can be observed that performance for SA was overall low compared to what could be expected from children aged 7-8 years: $\bar{x}=0.45$ for SA1 and $\bar{x}=0.225$ for SA2. We have two complementary explanations for this. Firstly, as discussed in Section \ref{childproced}, children had to read the tests on a screen, after which they had to type answers in open text fields. This is a challenging task by itself that relies on additional skills including language proficiency, conscientiousness, digital literacy, and more. Secondly, whereas `passing' originally only means that a child can work out where Sally will look (for the ball, or for Anne on her way to buy ice cream), we also asked for a motivation, which makes the test more demanding. For the SS, completed by the same group of children, we see the expected pattern that scores show a downward tendency as test items become increasingly difficult. The older group, aged 9-10, completed the IM. As discussed in Section \ref{IMres}, scores resonate with earlier work. Given that we see child performance not as the central phenomenon under observation in this paper, but rather as a reference for LLM performance, further discussion is outside our scope.

\section{Discussion}\label{discussion}
Summing up the results for the Sally-Anne tests, while it is less surprising that base-LLMs and smaller instruct-LLMs struggle with increasing test complexity and deviations, it is striking that second-order ToM immediately perturbs some large instruct-LLMs (e.g. PaLM2-chat), and that adding deviations from the original test formulations pushed performance of even the most competitive models down (e.g. GPT-4, GPT-3.5). This initially suggests that performance on ToM tasks does not generalize well beyond a few standard contexts in LLMs, in line with earlier work \citep{sap-etal-2022-neural, shapira2023clever, ullman2023large}.  

For the Strange Stories we saw that base-LLMs perform generally below child level. Most instruct-LLMs perform close to or above child level, particularly as items become more complex and child performance drops much more dramatically than LLM performance. Levels of deviation from the original test formulation seem to have made almost no impact for the SS, suggesting that the capacity to deal with non-literal language targeted by the Strange Stories test \textit{does} generalize to novel contexts. We conclude that instruct-LLMs are quite capable at interpreting non-literal language, a skill that in humans involves ToM. Since the training data of LLMs includes numerous books and fora, which are typically rich in irony, misunderstanding, jokes, sarcasm, and similar figures of speech, we tentatively suggest that LLMs are in general well-equipped to handle the sort of scenarios covered in the Strange Stories. This should in theory include base-LLMs, but it could be that their knowledge does not surface due to the test format, even after specialized prompting. Going one step further, we hypothesize that Sally-Ann is generally harder for LLMs given that this test relies less on a very specific sort of advanced language ability, but more on a type of behaviourally-situated reasoning that LLMs have limited access to during training \citep[see also][]{mahowald2023dissociating}.

The Imposing Memory test was the most challenging for both base- and instruct-LLMs. Since our version of this test was never published before, it constitutes another robustness test, which only GPT-4 as largest instruct-LLM seems to pass well.

\begin{figure}
    \centering
    \includegraphics[width=\linewidth]{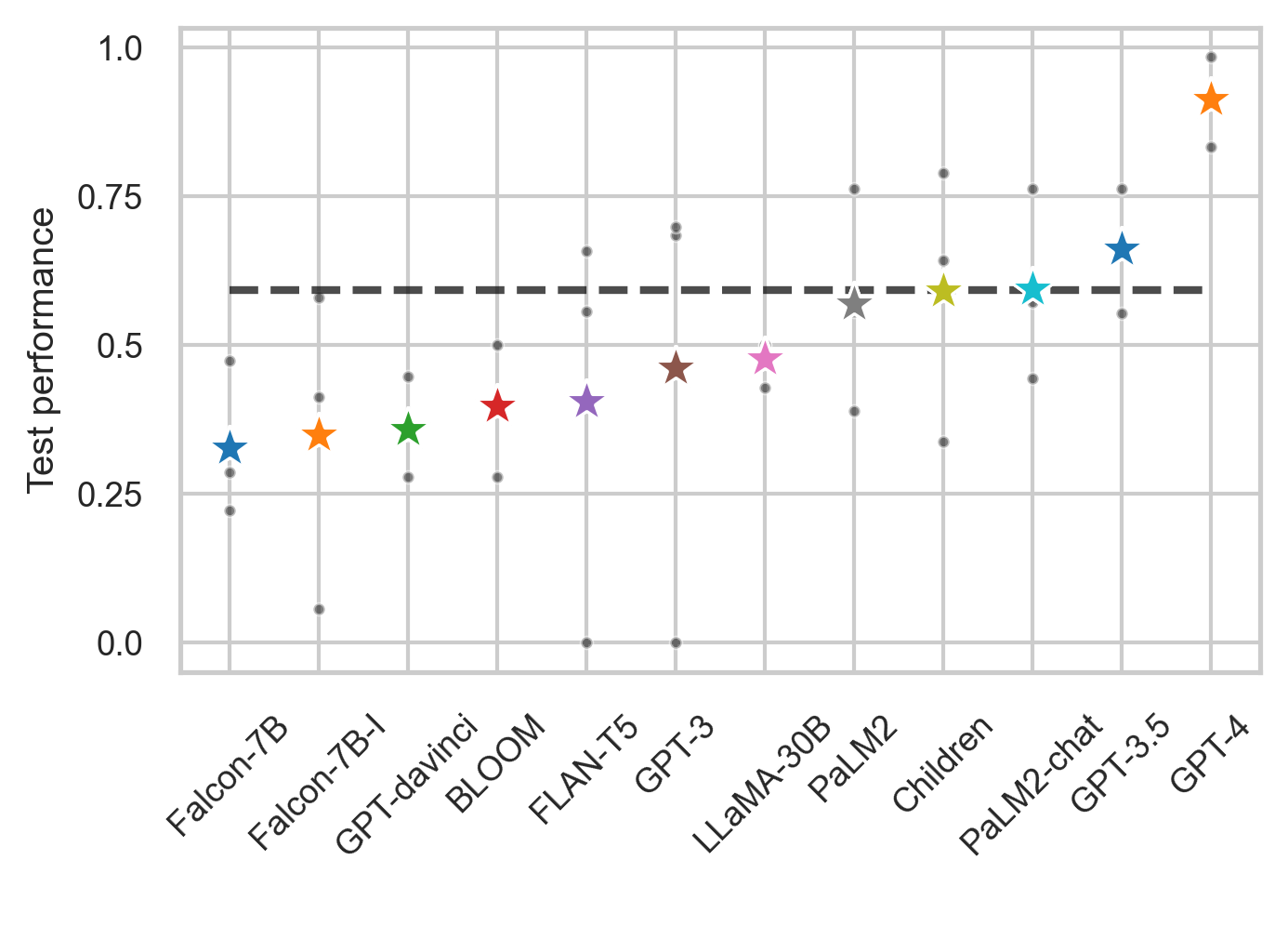}
    \caption{Grand mean performance (stars) of all mean test scores (dots) for children and LLMs.}
    \label{fig:overal_performance}
\end{figure}

The gap between base- and instruct-LLMs is best summarized in Figure \ref{fig:overal_performance}. Here we see that no base-LLM achieves child level: all LLMs approaching or exceeding child performance are larger instruct-LLMs. Our adapted prompts and insertion of correct answers for motivation questions did not make a difference. We suggest that another issue for base-LLMs, besides the prompt format, was prompt length. This was highest for IM, which can explain why they struggled most with this test. Prompt length, in relation to the models' varying context window sizes and ability to engage in what \citet{hagendorff2023intreasoning} call chain-of-thought reasoning, merits further research \citep[see also][]{liu2023lost}. We tested whether there was a difference between model performance on closed versus open questions across all three tasks, but found no signal: the models that struggled with closed questions were also those that performed low on open questions (for more details see Figure A on OSF).

Evidence is emerging that most LLM capacities are learned during self-supervised pre-training \citep{gudibande2023false, ye2023comprehensive}, which suggests that base-LLMs are essentially `complete' models. Yet instruction-tuning, even in small amounts \citep{zhou2023lima}, adds adherence to the desired interaction format and teaches LLMs, as it were, to apply their knowledge appropriately. We see a parallel between instruction-tuning and the role for \textit{rewarding cooperative communication} in human evolution and development. It has been argued extensively that human communication is fundamentally cooperative in that it relies on a basic ability and willingness to engage in mental coordination \citep[e.g][]{Verhagen2015coop, grice1975logic}. It is a key characteristic of the socio-cultural niche in which we evolved that, when growing up, we are constantly being rewarded for showing such willingness and cooperating with others to achieve successful communicative interactions \citep{tomasello2008origins}. Reversely, if we do not, we are being punished, explicitly or implicitly via increasing social exclusion \citep{davidbarrett2016language}. This brings us back to our context: instruction-tuning essentially rewards similar cooperative principles, but punishes the opposite, which may amount to an enhanced capacity for \textit{coordinating with an interaction partner's perspective}, in humans and LLMs alike. This is reflected in performance on ToM tasks, which are banking on this capacity too. 

Finally, we do not claim that LLMs that performed well also have ToM in the way that humans have it. Validity of cognitive tests such as those used in ToM research is a general issue \citep[e.g.][]{vanduijn2016}. Yet for humans ToM tests are validated `quick probes': decades of research have shown that proficiency on such tests \textit{correlates} with an array of real-world social and cognitive abilities \citep{beaudoin2020systematic}. For LLMs we are in a very early stage of figuring out what is entailed by proficon ToM tests: on the one hand it is impressive that some models show a degree of robust performance, without explicit training on ToM. On the other hand it remains an open question whether this amounts to any actual capacities in the social-cognitive domain, in which they are clearly very differently grounded (if at all) compared to humans. 

For future research we believe in the format of testing models that differ in other respects than just size, on a varied array of tasks, with multiple tests per test item, to gain further insight into the aspects that explain variability in performance. For this, more openness about architecture and training procedures of current and future LLMs is imperative. In addition, we believe to have contributed to the debate by benchmarking LLM results on child data, but more of this is needed. We had limited samples and age distributions, and tests were not presented in optimal ways (see Section \ref{childproced}). 

We emphasize that our results need to be seen within the time frame of late Spring 2023. The fast pace with which LLMs are currently released and, in some cases, updated, makes them a moving target. Moreover, there are indications that specific capacities of models from the GPT-family have declined over time, perhaps as a result of such updates; for example their ability to handle math problems and produce code \citep{chen2023chatgpt}. Future studies need to address how such developments impact the capacities assessed in this paper. 





\section{Conclusion}
We have shown that a majority of recent LLMs operate below performance of children aged 7-10 on three standardized tests relevant to ToM. Yet those that are largest in terms of parameters, and most heavily instruction-tuned, surpass children, with GPT-4 well above all other models, including more recent competitors like PaLM2-chat and PaLM2 (see Figure \ref{fig:overal_performance}). We have interpreted these findings by drawing a parallel between instruction-tuning and rewarding cooperative interaction in human evolution. We concede that researching the degree to which LLMs are capable of anything like thought in the human sense has only just begun, which leaves the field with exciting challenges ahead.

\section*{Acknowledgements}
This research was financed by the Dutch Research Council NWO (VI.Veni.191C.051). We are grateful to the children and their caregivers and teachers for participating in our research, and we thank Li Kloostra, Lola Vandame, and three anonymous reviewers for their help and constructive feedback.

\bibliography{anthology,custom}
\bibliographystyle{acl_natbib}




%

\end{document}